\documentclass[letterpaper]{article} % DO NOT CHANGE THIS
\usepackage{aaai2026}  % DO NOT CHANGE THIS
\usepackage{times}  % DO NOT CHANGE THIS
\usepackage{helvet}  % DO NOT CHANGE THIS
\usepackage{courier}  % DO NOT CHANGE THIS
\usepackage[hyphens]{url}  % DO NOT CHANGE THIS
\usepackage{graphicx} % DO NOT CHANGE THIS
\usepackage{natbib}  % DO NOT CHANGE THIS AND DO NOT ADD ANY OPTIONS TO IT
\usepackage{caption} % DO NOT CHANGE THIS AND DO NOT ADD ANY OPTIONS TO IT
\usepackage{algorithm}
\usepackage{algorithmic}

\usepackage{multirow}
\usepackage{booktabs}
\usepackage{breqn}
\usepackage{tabularx}
\usepackage{amssymb}
\usepackage{makecell}

\usepackage{newfloat}
\usepackage{listings}
\DeclareCaptionStyle{ruled}{labelfont=normalfont,labelsep=colon,strut=off} % DO NOT CHANGE THIS
\floatstyle{ruled}
\newfloat{listing}{tb}{lst}{}
\floatname{listing}{Listing}
\title{DSFedMed: Dual-Scale Federated Medical Image Segmentation via Mutual Distillation Between Foundation and Lightweight Models}
\author {
    Hanwen Zhang, 
    Qiaojin Shen, 
    Yuxi Liu, 
    Yuesheng Zhu, 
    Guibo Luo\thanks{Corresponding author. }
}
\affiliations {
    Guangdong Provincial Key Laboratory of Ultra High Definition Immersive Media Technology, \\ Shenzhen Graduate School, Peking University\\

    hanwen@stu.pku.edu.cn, qjshen25@stu.pku.edu.cn, yuxiliu@stu.pku.edu.cn, zhuys@pku.edu.cn, luogb@pku.edu.cn\\
}

\usepackage{bibentry}
\begin{document}

\maketitle

\begin{abstract}
Foundation Models (FMs) have demonstrated strong generalization across diverse vision tasks. However, their deployment in federated settings is hindered by high computational demands, substantial communication overhead, and significant inference costs. We propose DSFedMed, a dual-scale federated framework that enables mutual knowledge distillation between a centralized foundation model and lightweight client models for medical image segmentation. To support knowledge distillation, a set of high-quality medical images is generated to replace real public datasets, and a learnability-guided sample selection strategy is proposed to enhance efficiency and effectiveness in dual-scale distillation. This mutual distillation enables the foundation model to transfer general knowledge to lightweight clients, while also incorporating client-specific insights to refine the foundation model. Evaluations on five medical imaging segmentation datasets show that DSFedMed achieves an average 2 percent improvement in Dice score while reducing communication costs and inference time by nearly 90 percent compared to existing federated foundation model baselines. These results demonstrate significant efficiency gains and scalability for resource-limited federated deployments. 
\end{abstract}

% % Uncomment the following to link to your code, datasets, an extended version or similar.
% % You must keep this block between (not within) the abstract and the main body of the paper.
\begin{links}
    \link{Code}{https://github.com/LMIAPC/DSFedMed}
    % \link{Datasets}{https://aaai.org/example/datasets}
    % \link{Extended version}{https://aaai.org/example/extended-version}
\end{links}

\section{Introduce}

\begin{figure}[t]
\centering
\includegraphics[width=0.98\columnwidth]{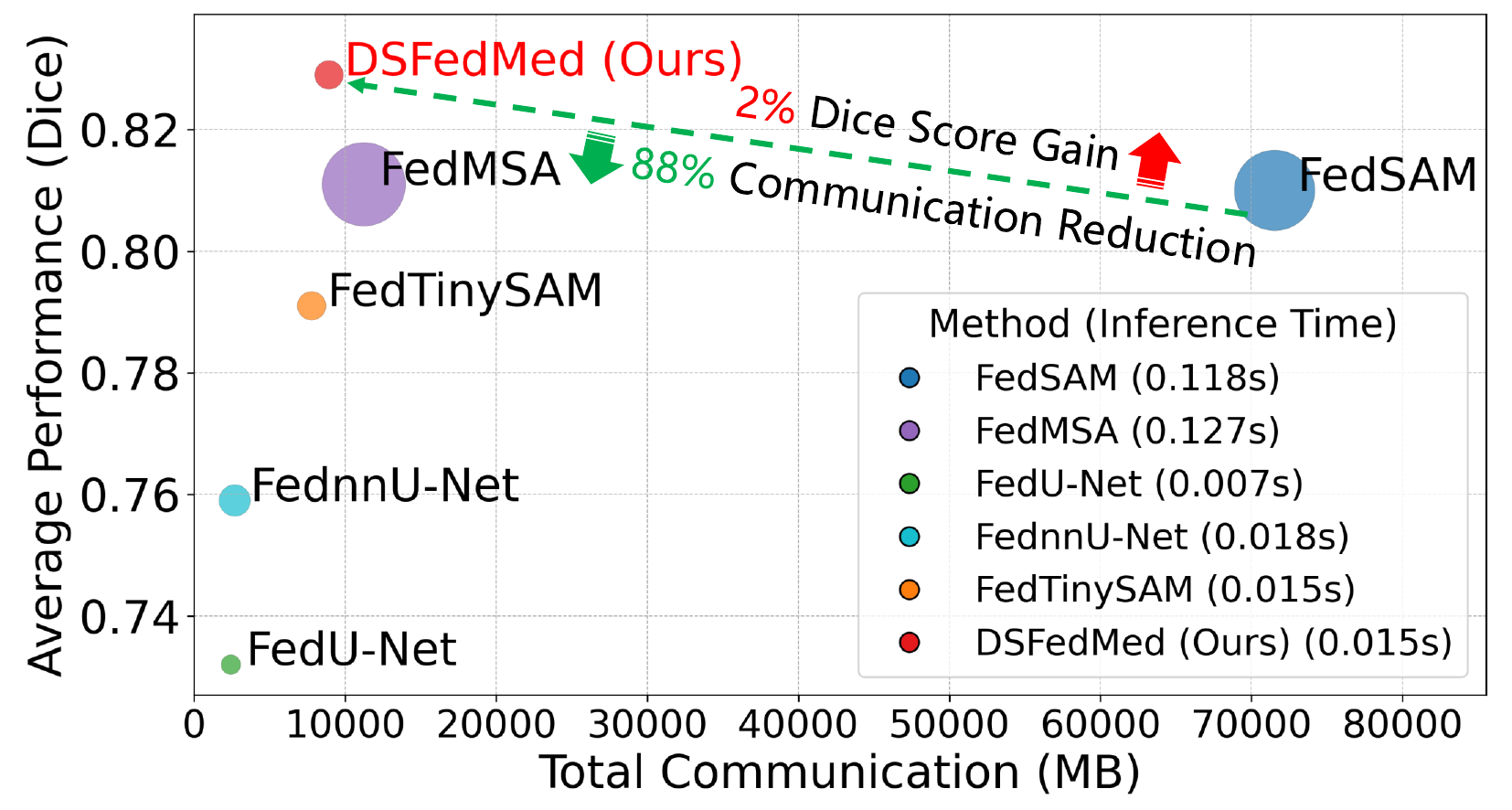}
\caption{Comparison of federated segmentation methods in terms of communication cost and Dice performance. Bubble size represents inference time. Foundation model-based methods (e.g., FedSAM) achieve high accuracy but suffer from heavy communication and inference overhead. Lightweight models (e.g., FedU-Net) are efficient but sacrifice accuracy. Our proposed method, DSFedMed, achieves a better balance, offering both high accuracy and low computational and communication costs.}
\end{figure}

\begin{figure*}[t]
\centering
\includegraphics[width=0.88\textwidth]{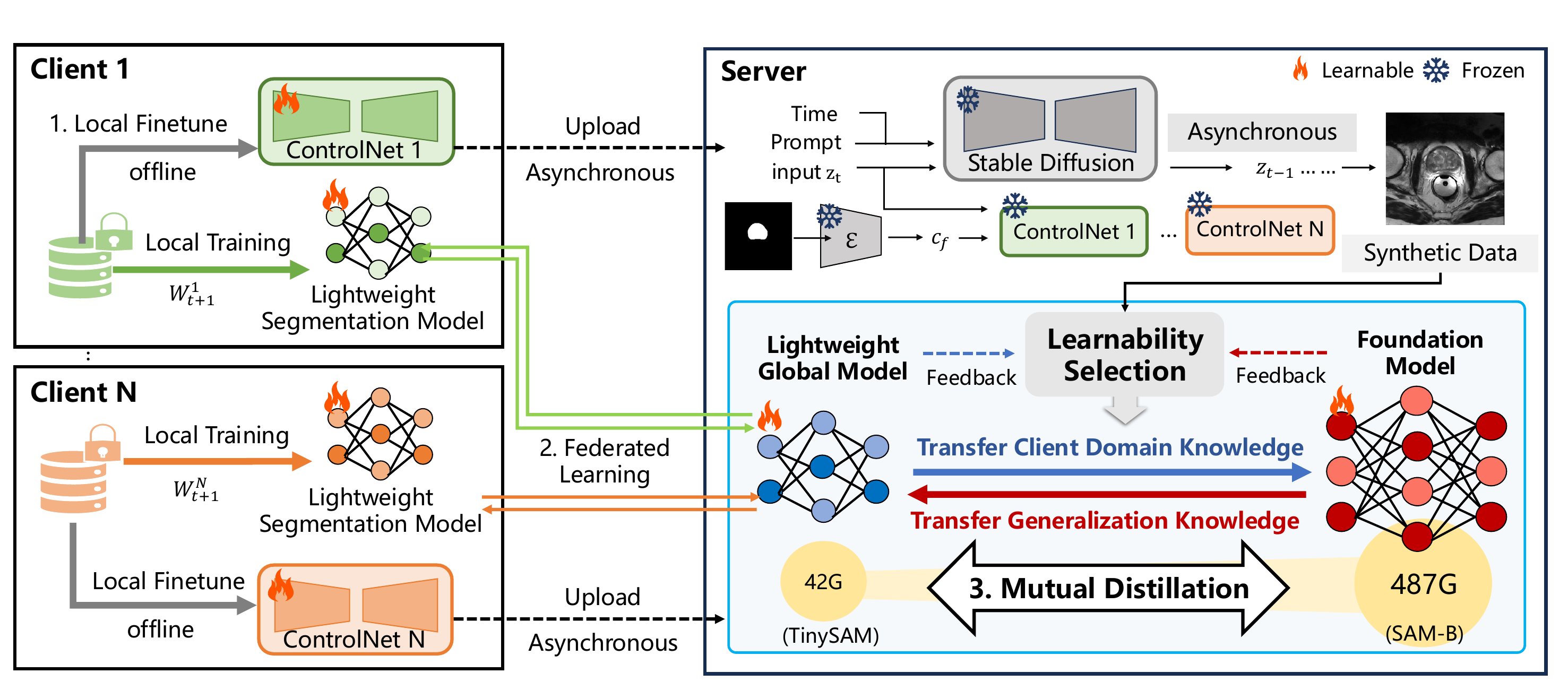}
\caption{The Overview of DSFedMed.}
\end{figure*}

The rise of Foundation Models (FMs) \cite{bommasani2021opportunities} has brought a paradigm shift to Artificial Intelligence (AI) \cite{zhuang2023foundation}. Models like Vision Transformers (ViTs) \cite{dosovitskiy2020image, steiner2021train} and the Segment Anything Model (SAM) \cite{kirillov2023segment} demonstrate strong generalization across diverse vision tasks. Pre-trained on large-scale datasets \cite{guo2022domain}, these models offer transferable representations with promising applications in real-world domains such as medical imaging, where robustness and adaptability are critical.

While general-purpose FMs benefit from abundant web-scale data, domain-specific FMs, particularly in sensitive fields like healthcare, require access to high-quality private data that cannot be easily centralized due to strict privacy regulations \cite{GDPR2016, CCPA2023} and institutional boundaries. This motivates the need for federated foundation models, where knowledge from distributed data silos can be aggregated without relying on direct access to local data.

Deploying foundation models directly in federated learning (FL) \cite{mcmahan2017communication, konevcny2016federated} presents significant challenges, especially in the medical imaging domain. Large foundation models require substantial computational resources and memory \cite{su2024fedra}, which often exceed the capabilities of client devices commonly used in clinical environments \cite{abbas2024federated}. Moreover, transmitting large model between clients and servers leads to heavy communication overhead, straining network bandwidth and increasing synchronization latency. Although parameter-efficient tuning methods like adapters can reduce training complexity to some extent \cite{malaviya2023reducing, cai2023efficient, chen2024feddat}, they do not solve the high inference cost and large memory requirement during deployment. These constraints are reflected in the practical trade-offs observed across federated segmentation methods. Figure 1 shows that existing approaches often prioritize either accuracy or efficiency, but rarely achieve both.

These practical limitations highlight the necessity for collaborative frameworks that integrate the strengths of large foundation models and lightweight models in a  federated setting. Specifically, lightweight models deployed on client devices enable efficient, real-time inference grounded in domain-specific knowledge, while powerful foundation models hosted on the server offer comprehensive global insights and guidance. This hybrid federated structure is especially beneficial in medical imaging scenarios, where client devices are severely resource-constrained and privacy regulations restrict data sharing \cite{ali2022federated}.

However, such dual-scale collaboration introduces several key challenges:

\begin{itemize}
    \item How to effectively leverage the generalization capabilities of large-scale foundation models within highly resource-constrained, decentralized environments remains an open question.
    \item Due to privacy concerns and regulatory constraints, real medical data cannot be exchanged or centralized. This raises the need to enable knowledge transfer without access to real public datasets. 
    \item Effective mutual knowledge distillation across heterogeneous model scales depends on selecting highly informative and reliable training samples, which makes learnability-guided sample selection crucial for efficient and accurate knowledge alignment.
\end{itemize}

In this paper, we propose \textbf{DSFedMed}, a dual-scale federated segmentation framework that addresses the challenges of deploying large foundation models in resource-constrained medical imaging environments. Specifically, our contributions:
\begin{itemize}
    \item A dual-scale federated knowledge transfer framework enabling collaboration between server-side foundation model and client-side lightweight model for medical image segmentation, with efficient client inference and communication for real-world deployment.
    \item A medical image generator that produces controllable and modality-adaptive samples aligned with the global data distribution based on ControlNet. 
    \item A learnability-guided mutual knowledge distillation mechanism supported by generated samples to efficiently identify and transfer the most informative knowledge between heterogeneous models.
\end{itemize}

Our method provides a practical and scalable solution for integrating FMs into FL systems, offering new possibilities for privacy-compliant large-model deployment in real-world decentralized environments. 

\begin{figure*}[t]
\centering
\includegraphics[width=0.88\textwidth]{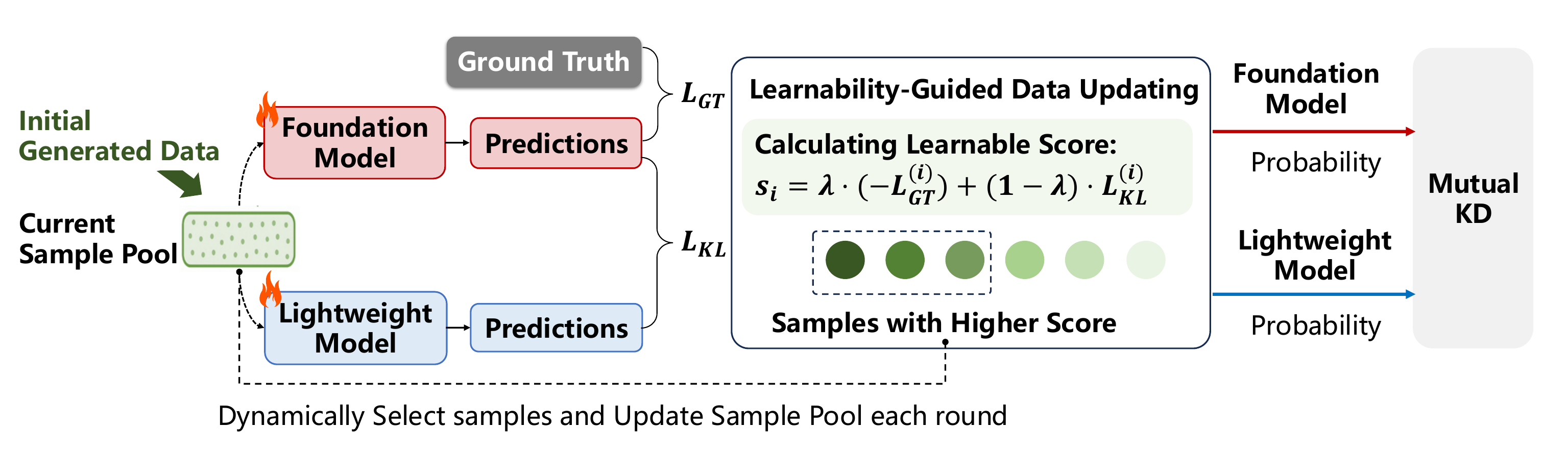}
\caption{Illustration of proposed learnability-guided mutual distillation.}
\end{figure*}

\section{Method}
\subsection{Overview}

We develop \textbf{DSFedMed}, a \textbf{\underline{D}}ual-\textbf{\underline{S}}cale \textbf{\underline{Fed}}erated \textbf{\underline{Med}}ical image segmentation framework that enables collaborative knowledge transfer between a centralized foundation model and client-side lightweight models under federated setting. An overview of the framework is shown in Figure 2, which consists of three main components: (1) Controllable medical image generation based on client-specific generator. (2) Federated learning and alignment of dual-scale segmentation models. (3) Learnability-guided mutual distillation using generated samples. 

In this framework, A lightweight segmentation model TinySAM \cite{shu2025tinysam}, is deployed on each client for efficient local training and personalization. Meanwhile, a foundation model SAM-B (ViT-B/16), is hosted on the central server as a generalist model with strong generalization capacity. For simplicity, we refer to SAM-B as SAM throughout the paper.

To bridge the knowledge gap between these heterogeneous models while reducing exposure to sensitive data, we propose a mutual federated distillation framework supported by high-quality generated data. Specifically, each client finetunes a ControlNet \cite{zhang2023adding} module, initialized from released mask-conditioning weights, while keeping the backbone diffusion model frozen. This enables controllable and modality-adaptive medical image generation from structural masks. The ControlNet weights are then uploaded to the server and used to generate a global set of image–mask pairs, which are consistent in client-specific modality characteristics while preserving the structural information. These generated samples enable learnability-guided mutual distillation between server-side SAM and client-side TinySAM. By scoring each sample based on supervision reliability and model disagreement, we dynamically select the most informative subset to enhance SAM's domain adaptation and TinySAM's generalization, enabling efficient dual-scale collaboration.

\subsection{Efficient Data Generation}

To enable distillation without accessing real datasets, we construct an offline synthetic data generation pipeline based on ControlNet and Stable Diffusion \cite{rombach2022high}. Building on the general insight from prior work \cite{jin2025fedwsiddfederatedslideimage} that synthetic data can support distillation, our framework allows clients to finetune only lightweight ControlNet modules using private image–mask pairs while freezing the diffusion backbone. This reduces training cost and memory usage. A low VRAM mode is also available for 8GB GPUs. The framework enables client-specific generation without altering the pre-trained diffusion model structure. 

Given a structural mask $m$ as condition $c$, ControlNet integrates this control into the diffusion process by extending the frozen diffusion network $F(x;\Phi)$ with a trainable branch $\Phi_C$, connected via zero-initialized 1×1 convolution layers $Z(\cdot;\Phi_{z1})$,$Z(\cdot;\Phi_{z2})$:
\begin{equation}
y_c=F(x;\Phi)+Z(F(x+Z(c_f;\Phi_{z1});\Phi_c);\Phi_{z2})
\end{equation}
Initially, $y_c=F(x;\Phi)$, preserving stable feature flow; during training, $\Phi_c$ learns to conditionally modulate generation via encoded condition $c_f=E(c)$, achieving fine structural alignment with high image fidelity. After offline training, each client uploads its personalized ControlNet parameters $\Phi_c^{(k)}$ to the server. The server then produces image-mask pairs from all clients' ControlNets by sampling local masks $m\sim\mathcal{M}_k$ for each client $k$:
\begin{equation}\tilde{D}=\bigcup_{k=1}^K\left\{(\tilde{x}_k,m)\mid\tilde{x}_k=g^{\mathrm{ctrl}}\left(m;\Phi_c^{(k)}\right),m\sim\mathcal{M}_k\right\}\end{equation}
This generated dataset enables knowledge distillation without relying on real data.

\subsection{Collaborative Training of Dual-Scale Model}
Each client $k\in\{1,...,K\}$ participates in the federated training of the lightweight TinySAM model $f_k^{\mathrm{Tiny}}$ on private data $\mathcal{D}_k^{\mathrm{real}}$ by minimizing segmentation loss $\mathcal{L}_{\sup}$:
\begin{equation}
\min_{\theta_{k}}\mathcal{L}_{\mathrm{sup}}(f_{k}^{\mathrm{Tiny}}(x_{k}),y_{k}),(x_{k},y_{k})\thicksim\mathcal{D}_{k}^{\mathrm{real}}
\end{equation}
After each local training iteration, clients upload their updated model parameters $\theta_k$ to the server. The server performs aggregation via federated averaging, weighting each client’s parameters by its dataset size:
\begin{equation}
\theta^{Tiny}=\sum_{k=1}^{K}\frac{n_{k}}{n}\theta_{k},n_{k}=|\mathcal{D}_{k}^{real}|~,~n=\sum_{k=1}^{K}n_{k}
\end{equation}
Meanwhile, the server uses the uploaded ControlNet to produce a global generated dataset $\tilde{D}$ as described above. The global SAM model $f^{\mathrm{SAM}}$ is finetuned on $\tilde{D}$:
\begin{equation}
\min_{\theta^{\mathrm{SAM}}}\mathcal{L}_{\mathrm{sup}}(f^{\mathrm{SAM}}(\tilde{x}),m)~,~(\tilde{x},m)\in\tilde{D}
\end{equation}
Although TinySAM and SAM are trained on real and generated data respectively, the client-specific ControlNet modules ensure consistent semantic styles across generated samples, thereby fostering a shared representation space that facilitates knowledge alignment between two model scales.

\subsection{Learnability-Guided Mutual Distillation}

To enable effective knowledge exchange between heterogeneous models, we propose a learnability-guided mutual distillation mechanism that iteratively selects and updates a dynamic sample pool from initially generated data, as illustrated in Figure 3. The process begins with an initial generated dataset $\tilde{D}=\{(x_i,m_i)\}_{i=1}^N$, where each image $x_i$ is generated from a structural mask $m_i$ as described above. Rather than using the entire dataset uniformly, we maintain a current sample pool that is continuously updated each round through a dynamic selection strategy guided by sample-wise informativeness. To evaluate each sample’s value for mutual distillation, we compute a learnability score $x_i$ by combining two complementary metrics: the ground truth loss (GT-loss) with respect to SAM's prediction, and the mutual KL divergence measuring the discrepancy between SAM and TinySAM outputs:
\begin{equation}
s_i=\lambda\cdot\left(-L_{GT}^{(i)}\right)+(1-\lambda)\cdot L_{KL}^{(i)}~,~\lambda\in[0,1]
\end{equation}
where:
\begin{equation}
\begin{split}
    \mathcal{L}_{GT}^{(i)}&=\ell(f_{SAM}(x_i),m_i),\\\mathcal{L}_{KL}^{(i)}
                &=D_{KL}\left(p_{SAM}(x_i)\parallel p_{\mathrm{TinySAM}}(x_i)\right) \\
                &+D_{KL}\left(p_{\mathrm{TinySAM}}(x_i)\parallel p_{SAM}(x_i)\right)
\end{split}
\end{equation}
Here, $L_{GT}$ ensures that the sample provides reliable SAM supervision, while $L_{KL}$ prioritizes samples where predictions diverge, indicating a greater potential for improvement between dual-scale models.

During training, we dynamically select samples based on updated scores using a fixed selection rate, refreshing the sample pool each round. This sample selection process can be viewed as an implicit mutual distillation, where the most informative examples guide the knowledge transfer between models:
\begin{equation}
\begin{split}
    \mathbb{E}_{x\sim a(x)} [ &D_{KL}\left(p_{SAM}(x)\parallel p_{\mathrm{TinySAM}}(x)\right) \\
    +&D_{KL}\left(p_{\mathrm{TinySAM}}(x)\parallel p_{SAM}(x)\right) ]
\end{split}
\end{equation}
where the sampling distribution $a(x)\propto\exp{(s_i)}$ biases towards samples that maximize mutual knowledge transfer, both SAM and TinySAM are fully trainable during this process. This learnability-guided scheme accelerates convergence by focusing on informative and reliable supervision.

Overall, our mutual distillation framework exploits a curated subset to transfer general representations from SAM to TinySAM, enhancing the lightweight model’s generalization, while injecting domain-specific knowledge from TinySAM back to SAM via disagreement samples. The example of learnability-guided selection is shown in Figure 4.

\begin{figure}[t]
\centering
\includegraphics[width=0.95\columnwidth]{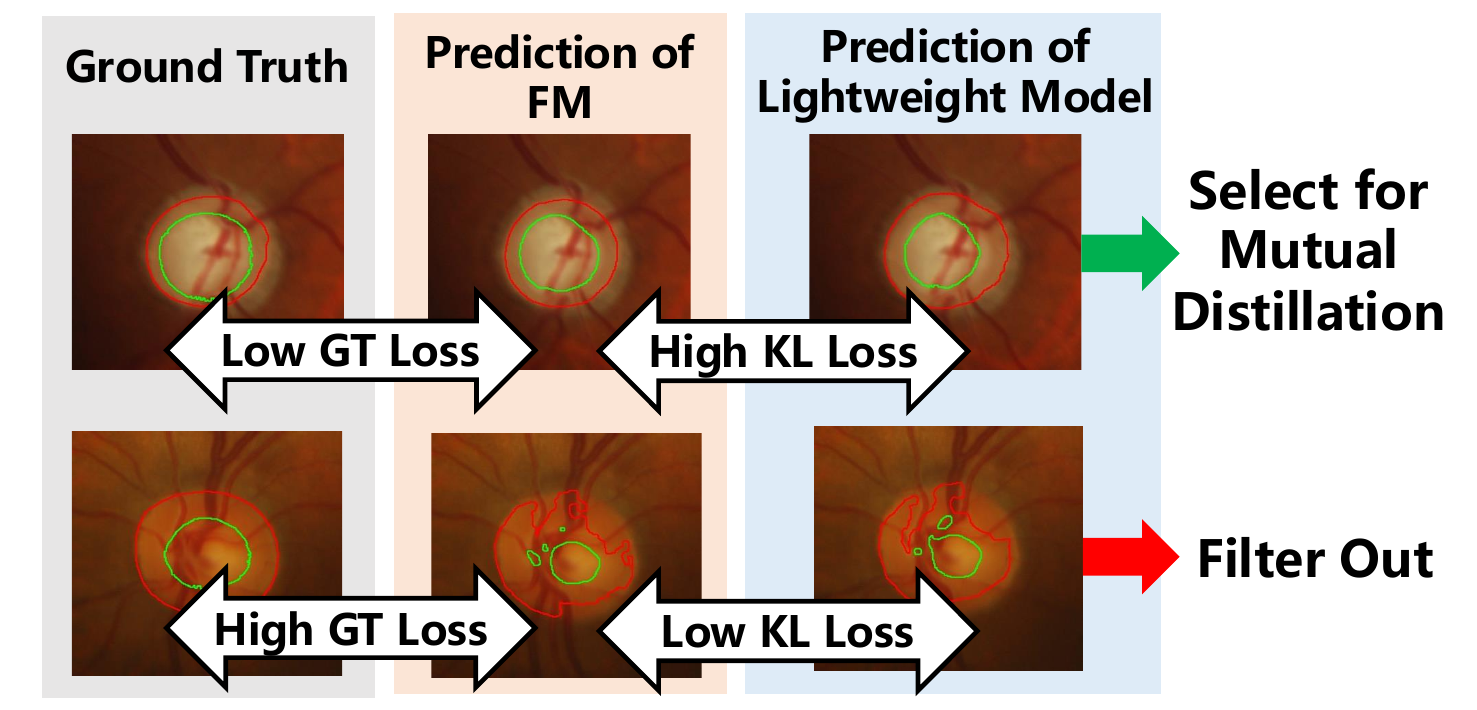}
\caption{Examples of Learnability-Guided Selection.}
\end{figure}

\section{Experiments}

\subsection{Datasets}

% We constructed five Non-IID federated datasets from public sources, covering diverse modalities and tasks: Fundus (retinal), Prostate (MRI), Nuclei (histology), ISIC (dermoscopy), and CHAOS (CT/MRI). Following the preprocessing pipeline of Medical SAM Adapter (MSA) \cite{wu2025medical}, all input images were resized to $1024\times1024$ to retain detail, and the corresponding segmentation masks are standardized to $256\times256$ resolution for efficiency.

We constructed five Non-IID federated datasets from public sources, covering diverse modalities and tasks. Following the preprocessing pipeline of MSA \cite{wu2025medical}, all input images were resized to $1024\times1024$ to retain detail, and the corresponding segmentation masks are standardized to $256\times256$ resolution for efficiency.

\begin{itemize}
    \item \textbf{Fundus:} Collected from independent medical centers and captured by different scanning devices \cite{sivaswamy2015comprehensive,fumero2011rim,orlando2020refuge}, targeting segmentation of the Optic Disc (OD) and Optic Cup (OC) .
    \item \textbf{Prostate Cancer:} Derived from publicly available MRI datasets collected across multiple medical centers \cite{litjens2014evaluation,lemaitre2015computer}.
    % \item \textbf{Prostate Cancer:} Derived from publicly available MRI datasets collected across multiple medical centers, including the NCI-ISBI 2013 dataset and other medical institutions \cite{litjens2014evaluation,lemaitre2015computer}.
    \item \textbf{Nuclei:} Sourced from several nuclei segmentation datasets \cite{gamper2019pannuke,gamper2020pannuke,kumar2017dataset,naylor2018segmentation,verma2021monusac2020}, with samples from different tissue types in the PanNuke dataset distributed across clients to introduce heterogeneity.
    \item \textbf{ISIC:} Derived from the ISIC 2017 dataset \cite{codella2018skin} for skin lesion segmentation. To introduce heterogeneity, the training set was partitioned into three subsets based on skin tone \cite{bevan2022detecting}, each assigned to a distinct client.
    \item \textbf{CHAOS:} Sourced from a classic benchmark providing paired CT and MRI data for liver segmentation \cite{kavur2021chaos}. Two modalities were allocated to separate clients to simulate cross-modality federated learning. 
    % Training and testing sets were separated by patient to avoid data overlap.
\end{itemize}

For the real-world multi-center Fundus, Prostate, and Nuclei datasets, we use leave-one-domain-out cross-validation, holding out each client as the test domain in turn while others train collaboratively. Since ISIC and CHAOS are not naturally multi-center, we manually split them and adopt a standard federated learning setup with all clients training concurrently and evaluation on an independent public test set.

\begin{table}[t]
\centering
\renewcommand{\arraystretch}{1.1} % 压缩行高
\small
\begin{tabular}{ll}
\hline
Type & Method \\
\hline
Centralized & SAM \\
\hline
\multirow{2}{*}{Federated Foundation Model} & FedSAM \\
~ &  FedMSA \\
\hline
\multirow{3}{*}{Federated Lightweight Segmentation Model} &  FedU-Net \\ 
~ & FednnU-Net \\
~ & FedTinySAM \\
\hline
\begin{tabular}[l]{@{}l@{}}Dual-Scale FL \\ (Server: FMs; Client: Lightweight Models) \end{tabular} & DSFedMed \\
\hline
\end{tabular}
\normalsize
\caption{Categorization of compared baselines. }
\end{table}

\begin{table*}[t]
\centering
\renewcommand{\arraystretch}{1.1}
\small
\begin{tabular}{c cc cc cc cc cc cc}
\hline
Datasets & \multicolumn{2}{c}{Fundus-OD} & \multicolumn{2}{c}{Fundus-OC} & \multicolumn{2}{c}{Prostate} & \multicolumn{2}{c}{Nuclei} & \multicolumn{2}{c}{ISIC} & \multicolumn{2}{c}{CHAOS} \\
\hline
Method & Dice & IoU & Dice & IoU & Dice & IoU & Dice & IoU & Dice & IoU & Dice & IoU \\
\hline
SAM (ceiling) & 0.914 & 0.850 & \underline{0.782} & \underline{0.673} & 0.801 & 0.762 & 0.617 & 0.467 & 0.840 & 0.755 & \textbf{0.938} & \textbf{0.895} \\
\hline
FedSAM & \underline{0.926} & \underline{0.869} & 0.771 & 0.661 & \textbf{0.814} & \textbf{0.781} & \underline{0.623} & \underline{0.480} & 0.822 & 0.735 & 0.902 & 0.848\\
FedMSA & 0.926 & 0.868 & 0.764 & 0.647 & 0.799 & 0.764 & 0.617 & 0.472 & \underline{0.840} & \underline{0.755} & 0.917 & 0.864 \\
\hline
FedU-Net & 0.896 & 0.820 & 0.733 & 0.600 & 0.517 & 0.490 & 0.590 & 0.435 & 0.797 & 0.708 & 0.860 & 0.788 \\
FednnU-Net & 0.907 & 0.837 & 0.743 & 0.614 & 0.590 & 0.516 & 0.603 & 0.450 & 0.819 & 0.726 & 0.894 & 0.843 \\
FedTinySAM & 0.917 & 0.852 & 0.738 & 0.620 & 0.756 & 0.723 & 0.620 & 0.473 & 0.832 & 0.747 & 0.883 & 0.831 \\
\hline
DSFedMed & \textbf{0.949} & \textbf{0.906} & \textbf{0.797} & \textbf{0.686} & \underline{0.813} & \underline{0.777} & \textbf{0.644} & \textbf{0.498} & \textbf{0.842} & \textbf{0.758} & \underline{0.927} & \underline{0.876} \\
\hline
\end{tabular}
\normalsize
\caption{Segmentation results of different methods on five medical datasets. The best results are highlighted in bold and the suboptimal ones are underlined.} 
\end{table*}

\begin{figure*}[t]
\centering
\includegraphics[width=0.88\textwidth]{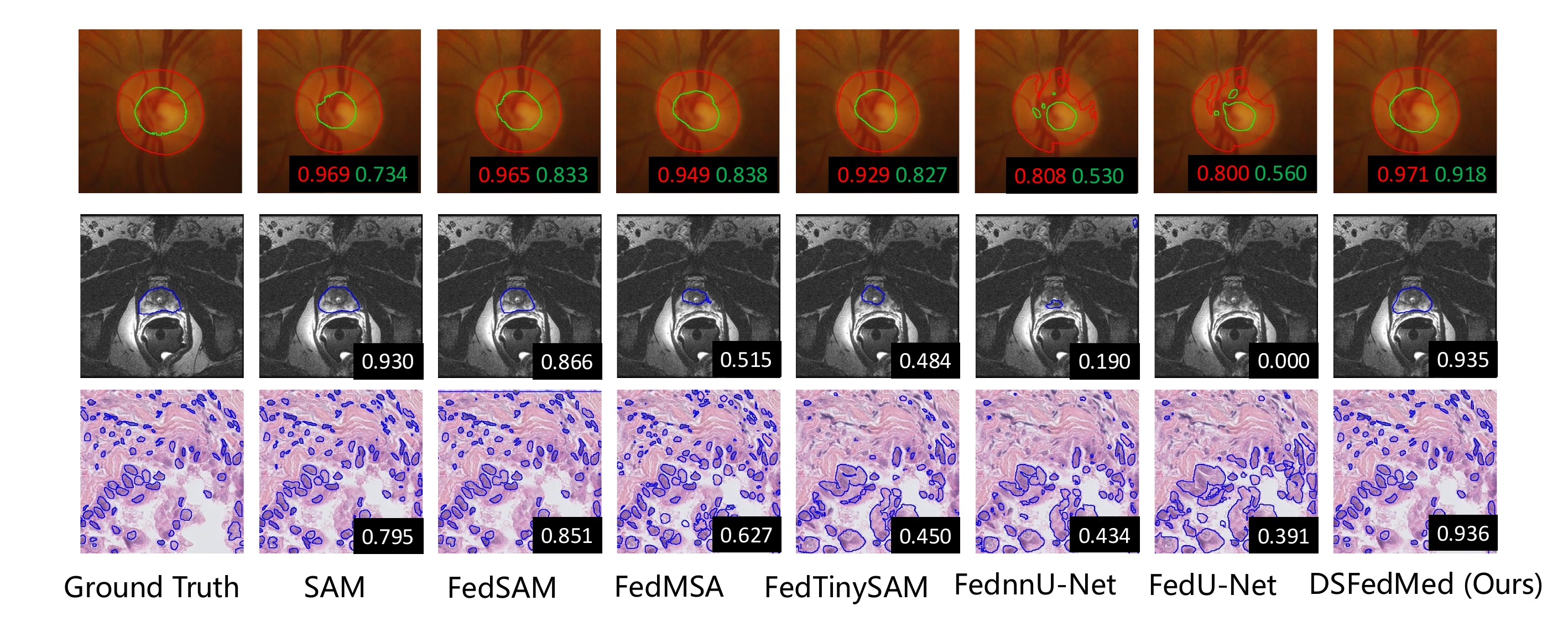}
\caption{The example of datasets and the comparison in the results of SAM, FedSAM, FedMSA, FedU-Net, FednnU-Net, FedTinySAM and DSFedMed. The average Dice of the method on the dataset is marked in the bottom right corner.}
\end{figure*}

\begin{figure}[t]
\centering
\includegraphics[width=0.98\columnwidth]{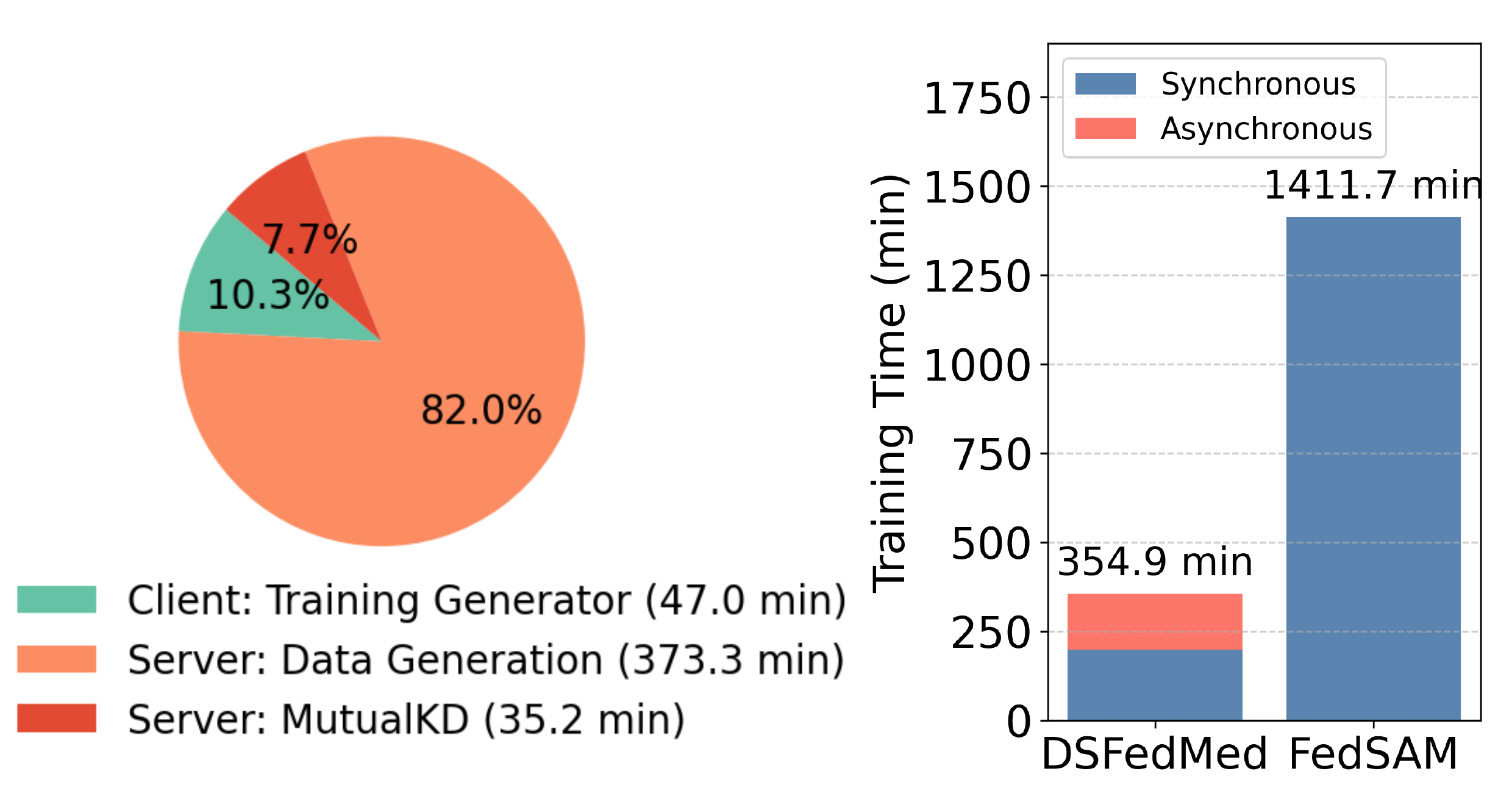}
\caption{Asynchronous training design of DSFedMed. (Left: The distribution of asynchronous training time. Right: Extra training time of DSFedMed and FedSAM for adding a new client.)}
\end{figure}

\subsection{Compared Baselines}

As shown in Table 1, we compare DSFedMed with the following baselines. Unless otherwise specified, both SAM and MSA are implemented with ViT-B/16 \cite{dosovitskiy2020image, steiner2021train}. When using the larger variant ViT-L/16, we denote the models as SAM-L and MSA-L.

\begin{itemize}
    \item \textbf{Centralized,} in which the server’s foundation model SAM is finetuned using the datasets combining private datasets of involved clients and the public dataset.
    \item \textbf{Federated Foundation Model,} on each client we adopt FedSAM to finetune the full SAM and FedMSA \cite{liu2024fedfms} to efficiently finetune the MSA adapter and decoder.    
    \item \textbf{Lightweight Federated Learning Baselines,} we adopt FedU-Net, FednnU-Net and FedTinySAM as our federated segmentation baselines. Both of them are implemented under the FedAvg framework \cite{mcmahan2017communication}. U-Net \cite{ronneberger2015u} is a widely used architecture for biomedical image segmentation, while nnU-Net \cite{isensee2018nnu} is a more robust extension of U-Net. TinySAM is a lightweight variant of SAM designed for edge efficiency.
\end{itemize}

\subsection{Efficiency Analysis}

\begin{table*}[t]
\centering
\renewcommand{\arraystretch}{1.1}
\small
\begin{tabular}{
  p{2cm}<{\raggedright}
  p{2cm}<{\raggedright}
  p{2cm}<{\raggedright}
  p{2cm}<{\raggedright}
  p{2.5cm}<{\raggedright}
  p{2cm}<{\raggedright}
  p{2cm}<{\raggedright}
}
\hline
Method & \begin{tabular}[c]{@{}l@{}}Inference\\ Model\end{tabular} & \begin{tabular}[c]{@{}l@{}} Communication\\ Overhead\end{tabular} & \begin{tabular}[c]{@{}l@{}}Inference Time\\(per image)\end{tabular} & \begin{tabular}[c]{@{}l@{}}Synchronous\\ Training Time\end{tabular} & \begin{tabular}[c]{@{}l@{}}Asynchronous\\ Training Time\end{tabular} & \begin{tabular}[c]{@{}l@{}}Average\\ Dice Score\end{tabular} \\
\hline
FedSAM & SAM-B & 71,538 MB & 0.118 s & 1,411.67 min & / & 0.810 \\
FedMSA & MSA-B & 11,230 MB & 0.127 s & 1,160.00 min & / & 0.811 \\
\hline
FedU-Net & U-Net & 2,430 MB & 0.007 s & 59.50 min & / & 0.732 \\
FednnU-Net & nnU-Net & 2,690 MB & 0.018 s & 345.68 min & / & 0.759 \\
FedTinySAM & TinySAM & 7,770 MB & 0.015 s & 198.02 min & / & 0.791 \\
\hline
DSFedMed & TinySAM & 8,920 MB & 0.015 s & 198.02 min  & 455.5 min  & 0.829 \\
\hline
\end{tabular}
\normalsize
\caption{Efficiency analysis of FL baselines and DSFedMed. }
\end{table*}

\subsection{Segmentation Result}

% Across all tasks and clients, DSFedMed consistently outperforms existing lightweight FL methods, including FedU-Net, FednnU-Net, and FedTinySAM. DSFedMed achieves an average Dice improvement of 3.8\% across five datasets compared to FedTinySAM: +3.2\% on Fundus-OD, +5.9\% on Fundus-OC, +5.7\% on Prostate, +2.4\% on Nuclei, +1.0\% on ISIC, and +4.4\% on CHAOS, respectively. 

Across all tasks and clients, DSFedMed consistently outperforms lightweight FL methods. DSFedMed achieves an average Dice improvement of 3.8\% across five datasets compared to FedTinySAM. Moreover, DSFedMed outperforms not only lightweight models but also strong foundation model baselines: it achieves a 1.4\% gain over centralized SAM, and surpasses FedSAM and the FedMSA by 1.9\% each. By leveraging high-quality generated data to facilitate mutual distillation, DSFedMed bridges the gap with foundation models and even exceeds their results in federated settings. We provide the results of the experiment in Table 2 and show the visualization results in Figure 5.

% Moreover, DSFedMed outperforms not only lightweight models but also strong foundation model baselines: it achieves a 1.4\% gain over centralized SAM, and surpasses FedSAM and the PEFT-based FedMSA by 1.9\% each. By leveraging high-quality generated data to facilitate mutual distillation, DSFedMed bridges the performance gap with foundation models and even exceeds their results in federated settings. We provide the results of the experiment in Table 2 and show the visualization results in Figure 5.

\begin{figure*}[t]
\centering
\includegraphics[width=0.90\textwidth]{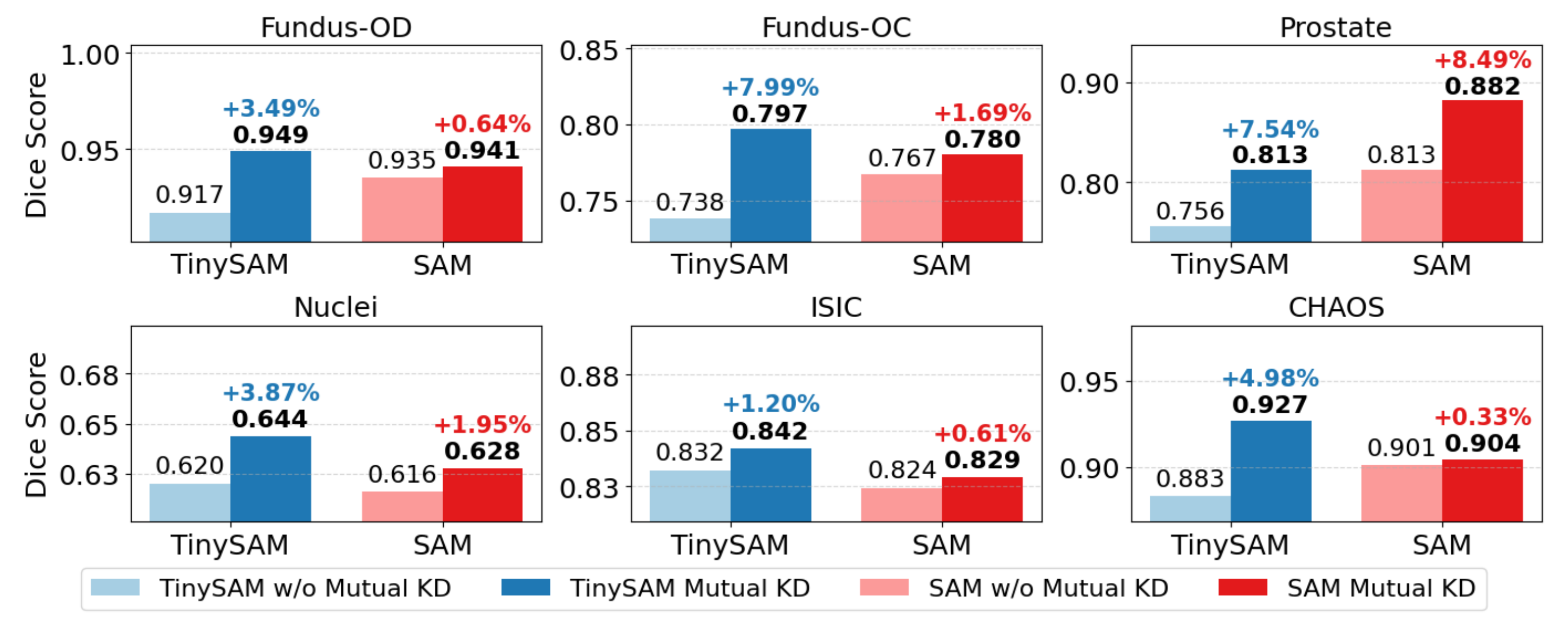}
\caption{Visualization of mutual enhancement across dual-scale models.}
\end{figure*}

\begin{table*}[t]
\centering
\small
\renewcommand{\arraystretch}{1.1} % 压缩行高
\begin{tabular}{p{0.8cm}<{\centering}p{1cm}<{\centering} cc cc cc cc cc cc c}
\hline
\multicolumn{2}{c}{Method} & \multicolumn{2}{c}{Fundus-OD} & \multicolumn{2}{c}{Fundus-OC} & \multicolumn{2}{c}{Prostate} & \multicolumn{2}{c}{Nuclei} & \multicolumn{2}{c}{ISIC} & \multicolumn{2}{c}{Chaos} & \multirow{2}{*}{\begin{tabular}[c]{@{}c@{}}~\\ time(s) \end{tabular}} \\
\cline{1-14}
\begin{tabular}[c]{@{}c@{}}Mutual\\ KD\end{tabular} & \begin{tabular}[c]{@{}c@{}}LG\\ Selection \end{tabular} & Dice & IOU & Dice & IOU & Dice & IOU & Dice & IOU & Dice & IOU & Dice & IOU & ~ \\
\hline
~ & ~ & 0.944  & 0.896  & 0.778  & 0.662  & 0.780  & 0.741  & 0.627  & 0.485  & 0.823  & 0.732  & 0.893  & 0.828 & 399.010 \\
\checkmark & ~ & 0.946  & 0.900  & 0.781  & 0.656  & 0.791  & 0.754  & 0.634  & 0.488  & 0.828  & 0.738  & 0.916  & 0.869 & 399.349 \\
~ & \checkmark & 0.945  & 0.900  & 0.788  & 0.674  & 0.794  & 0.764  & 0.636  & 0.491  & 0.839  & 0.752  & 0.911  & 0.852 & 209.827 \\
\checkmark & \checkmark & \textbf{0.949}  & \textbf{0.906}  & \textbf{0.797}  & \textbf{0.686}  & \textbf{0.813}  & \textbf{0.777}  & \textbf{0.644}  & \textbf{0.498}  & \textbf{0.842}  & \textbf{0.758}  & \textbf{0.927}  & \textbf{0.876} & 211.113 \\
\hline
\end{tabular}
\normalsize
\caption{Ablation study of key components in the proposed method. The best results are highlighted in bold. }
\end{table*}

As shown in Table 3, we compare federated learning methods in terms of total communication overhead, inference latency, training time, and segmentation performance. Communication overhead is computed cumulatively over 100 communication rounds, and training time is reported on the Prostate dataset.

FedSAM achieves a Dice score of 0.810 but is constrained by high communication cost, long training time, and slow inference. FedMSA improves efficiency via adapter-based tuning but incurs additional inference cost. Lightweight methods (e.g., FedU-Net, FednnU-Net) offer fast and efficient deployment but lack sufficient accuracy.

In contrast, DSFedMed strikes an optimal balance between segmentation accuracy and deployment efficiency, outperforming competing methods comprehensively. Compared to FedSAM, it reduces communication cost by a substantial 88\% while boosting the Dice score by 2\%, achieving highest accuracy of 0.829. Additionally, DSFedMed maintains a low inference time of 0.015 seconds, demonstrating both high accuracy and practical viability. 

The asynchronous training phase includes client-side generator training and server-side data generation and mutual knowledge distillation, as shown in Figure 6. Most asynchronous computations, including data generation and knowledge distillation, occur on the server without clients needing to stay online. This design enables new clients to join without full retraining. The server-side processing is efficient and can be performed on a single GPU, such as an NVIDIA A800 or A6000.

\subsection{Mutual Enhancement of Dual-Scale Model}

To assess the effectiveness of our dual-scale distillation framework, we compare the performance of both the lightweight model (TinySAM) and the foundation model (SAM) with and without mutual distillation. Experiments across five datasets demonstrate that mutual distillation consistently enhances both model scales as shown in Figure 7.

TinySAM achieves notable improvements, especially under domain shift, with Dice scores increasing on datasets like Prostate (+5.7\%) and Fundus-OC (+5.9\%). Performance gains are also observed on ISIC and CHAOS, confirming its enhanced generalization. Meanwhile, SAM improves on client-specific data, such as Prostate and Fundus-OD, indicating better domain adaptation.

These results demonstrate effective mutual knowledge exchange: SAM provides semantic priors to guide TinySAM, while TinySAM offers localized feedback to refine SAM.

\subsection{Ablation Studies}
% We conducted an ablation study to evaluate the contributions of Mutual Knowledge Distillation (Mutual KD) and Learnability-Guided sample selection (LG selection) as shown in Table 4. Here, Mutual KD replaces unidirectional distillation from server to client and consistently improves segmentation performance across five datasets, highlighting the advantage of bidirectional knowledge exchange for dual-scale adaptation. LG selection alone further enhances accuracy while reducing training time by nearly 50\%, demonstrating its effectiveness in identifying informative samples. When combined, both modules achieve the highest Dice and IoU scores with training efficiency comparable to LG selection alone, confirming their complementary roles in boosting both performance and efficiency within our collaborative knowledge transfer framework.

We conducted an ablation study to evaluate the contributions of Mutual Knowledge Distillation (Mutual KD) and Learnability-Guided sample selection (LG selection) as shown in Table 4. Replacing unidirectional distillation, Mutual KD consistently improves segmentation across all five datasets, highlighting the advantage of bidirectional knowledge exchange. LG selection alone boosts accuracy and cuts training time by nearly 50\%. Their combination achieves the highest Dice and IoU scores with training efficiency matching LG selection alone, confirming their complementary roles in our framework.

\subsection{Performance Across Different Model Scales}
The result in Table 5 shows that the proposed Dual-Scale Federated Learning framework consistently improves segmentation performance across combinations of server and client models on Fundus dataset. We evaluate with SAM-B (ViT-B/16) and Medical SAM Adapter-L (MSA-L, ViT-L/16) as server-side models, and U-Net and TinySAM as lightweight client models, demonstrating the framework’s general applicability across model scales.

\subsection{Sensitivity to Hyperparameters}
Experimental results show that sample selection consistently outperforms the baseline using all samples (selection rate = 1.0), achieving stable improved accuracy across selection rates from 0.2 to 0.8 while reducing training cost. Varying the weighting parameter \(\lambda\) between 0.1 and 0.9 induces only modest fluctuations in segmentation performance, with optimal outcomes observed near \(\lambda = 0.5\). These findings confirm the effectiveness and robustness of our framework. Figure 8 summarizes the results.

\begin{table}[t]
\centering
\renewcommand{\arraystretch}{1.1}
\small
\begin{tabular}{p{3.2cm}cccc}
\hline
\multirow{2}{*}{\shortstack[l]{Dual-Scale FL\\(Server / Client)}} & \multicolumn{2}{c}{Fundus-OD} & \multicolumn{2}{c}{Fundus-OC} \\
\cline{2-5}
& Dice & IoU & Dice & IoU \\
\hline
SAM-B / U-Net       & 0.940 & 0.891 & 0.739 & 0.613 \\
SAM-B / TinySAM     & 0.949 & 0.906 & 0.797 & 0.686 \\
MSA-L / U-Net       & 0.934 & 0.882 & 0.781 & 0.661 \\
MSA-L / TinySAM     & 0.944 & 0.897 & 0.795 & 0.681 \\
\hline
\end{tabular}
\normalsize
\caption{Performance of proposed Dual-Scale FL framework with different server/client model scales. SAM-B and MSA-L denote Base and Large variants, respectively.}
\label{tab:dualscale-performance}
\end{table}

\begin{table}[t]
\centering
\renewcommand{\arraystretch}{1.1} % 压缩行高
\small
% \begin{tabular}{ c p{0.6cm}<{\centering} p{0.6cm}<{\centering} p{0.6cm}<{\centering} p{0.6cm}<{\centering} p{0.6cm}<{\centering} p{0.6cm}<{\centering}}
\begin{tabular}{c cp{0.5cm}<{\centering}cp{0.5cm}<{\centering}cp{0.5cm}<{\centering}}
\hline
Datasets & \multicolumn{2}{c}{Fundus} & \multicolumn{2}{c}{Prostate} & \multicolumn{2}{c}{Nuclei} \\
\hline
Method & FID & FLD & FID & FLD & FID & FLD \\
\hline
Pix2Pix & 77.19 & 8.27 & 143.96 & 6.87 & 172.12 & 9.82 \\
DDPM & 34.86 & 9.33 &86.57 & 4.96 & 156.65 & 8.22 \\
ControlNet & \textbf{7.83} & \textbf{2.98} & \textbf{62.77} & \textbf{1.57} & \textbf{148.73} & \textbf{4.64} \\
\hline
\end{tabular}
\normalsize
\caption{Quantitative evaluation of the generated data. The best results are highlighted in bold.}
\end{table}

\begin{figure}[t]
\centering
\includegraphics[width=0.98\columnwidth]{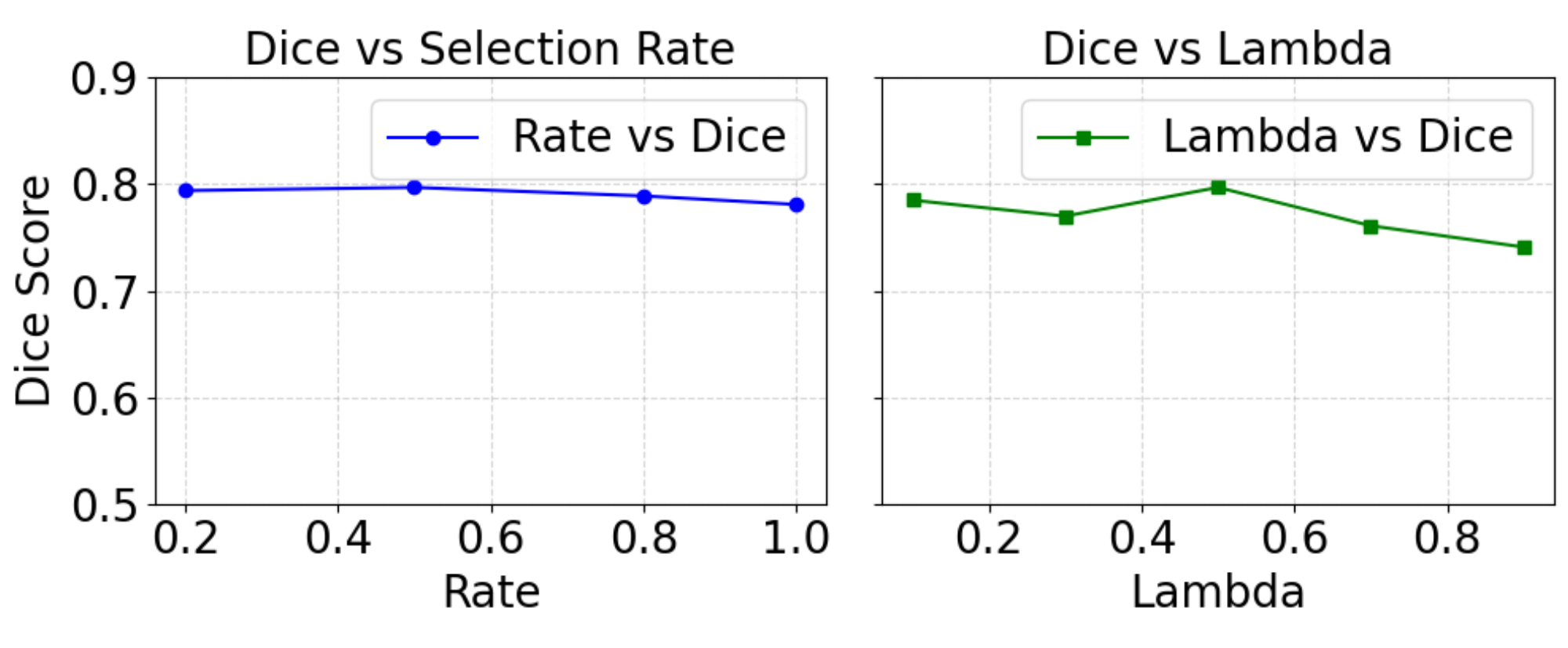}
\caption{Sensitivity to hyperparameters.}
\end{figure}

\subsection{Quality of Generated Data}

% In terms of generated data, we measured and evaluated three methods: pix2pix\cite{isola2017image}, DDPM\cite{ho2020denoising}, and ControlNet\cite{zhang2023adding}. To assess the quality of synthetic data, we adopt the Fréchet Inception Distance (FID) \cite{heusel2017gans} and the Feature Likelihood Difference (FLD) \cite{jiralerspong2023feature}. FID evaluates similarity of feature distributions, while FLD uses density estimation to measure novelty, realism, and diversity. Among the evaluated methods, ControlNet demonstrates superior performance, achieving the best balance of precision and controllability in synthetic image generation. Results are shown in Table 6. 

In terms of generated data, we measured and evaluated three methods: pix2pix\cite{isola2017image}, DDPM\cite{ho2020denoising}, and ControlNet\cite{zhang2023adding}, using Fréchet Inception Distance (FID) \cite{heusel2017gans} and the Feature Likelihood Difference (FLD) \cite{jiralerspong2023feature} to assess the quality of synthetic data. FID measures feature distribution similarity, while FLD assesses novelty, realism, and diversity via density estimation. ControlNet achieves the best balance of precision and controllability, outperforming the others. Results are in Table 6.

\section{Conclusion}
In this work, we proposed DSFedMed, a Dual-Scale Federated learning framework that enables collaboration between a centralized foundation model and distributed lightweight client models for Medical image segmentation. By integrating an efficient data generator and a learnability-guided sample selection strategy, our approach facilitates efficient mutual knowledge distillation without requiring access to real medical data or direct communication of foundation model.

Our results demonstrate the practicality and scalability of DSFedMed in resource-constrained federated settings, offering a viable path toward deploying foundation models in privacy-sensitive domains such as healthcare. The method balances global semantic generalization and local domain adaptability, addressing both performance and deployment limitations inherent in medical federated learning.

Nevertheless, several open challenges remain. These include accelerating data generation stage and validating the framework in real clinical settings. Future work may also extend DSFedMed to multimodal scenarios.

\section{Acknowledgments}
This work is supported by National Key Research and Development Program of China (2024YFE0203100), Guangdong Provincial Key Laboratory of Ultra
High Definition Immersive Media Technology (Grant No. 2024B1212010006), Shenzhen
Science and Technology Program (JCYJ20230807120800001), and the project supplementary funding of National Innovation 2030 Major S\&T Project of China (2020AAA0104203).

\bibliography{aaai2026}

\end{document}